\documentclass[10pt,twocolumn,letterpaper]{article}

\usepackage{cvpr}              

\usepackage[dvipsnames]{xcolor}

\definecolor{cvprblue}{rgb}{0.21,0.49,0.74}
\usepackage[pagebackref,breaklinks,colorlinks,citecolor=cvprblue]{hyperref}
\usepackage{amsmath}

\def\paperID{1811} 
\def\confName{CVPR}
\def\confYear{2024}

 \title{GaussianVTON: 3D Human Virtual Try-ON via Multi-Stage Gaussian Splatting Editing with Image Prompting}

\author{Haodong Chen$^1$ \and Yongle Huang$^1$  \and Haojian Huang$^2$ \and Xiangsheng Ge$^1$ \and 
Dian Shao$^1$\footnotemark[2]\\
\and
$^1$Northwestern Polytechnical University, $^2$The University of Hong Kong\\
}
 
\begin{document}
\maketitle
\renewcommand{\thefootnote}{\fnsymbol{footnote}} 
\footnotetext[2]{Corresponding authors.}
\begin{abstract}

The increasing prominence of e-commerce has underscored the importance of Virtual Try-On (VTON). However, previous studies predominantly focus on the 2D realm and rely heavily on extensive data for training. Research on  3D VTON primarily centers on garment-body shape compatibility, a topic extensively covered in 2D VTON. Thanks to advances in 3D scene editing, a 2D diffusion model has now been adapted for 3D editing via multi-viewpoint editing. In this work, we propose GaussianVTON, an innovative 3D VTON pipeline integrating Gaussian Splatting (GS) editing with 2D VTON. To facilitate a seamless transition from 2D to 3D VTON, we propose, for the first time, the use of only images as editing prompts for 3D editing. To further address issues, \textit{e.g.}, face blurring, garment inaccuracy, and degraded viewpoint quality during editing, we devise a three-stage refinement strategy to gradually mitigate potential issues. Furthermore, we introduce a new editing strategy termed Edit Recall Reconstruction (ERR) to tackle the limitations of previous editing strategies in leading to complex geometric changes. Our comprehensive experiments demonstrate the superiority of GaussianVTON, offering a novel perspective on 3D VTON while also establishing a novel starting point for image-prompting 3D scene editing. Project page: \textcolor{magenta}{\url{https://haroldchen19.github.io/gsvton/}}

\end{abstract}
   
\section*{1. Introduction}
\addcontentsline{toc}{section}{Introduction}

\begin{figure}
    \centering
    \small
   \includegraphics[width=\linewidth]{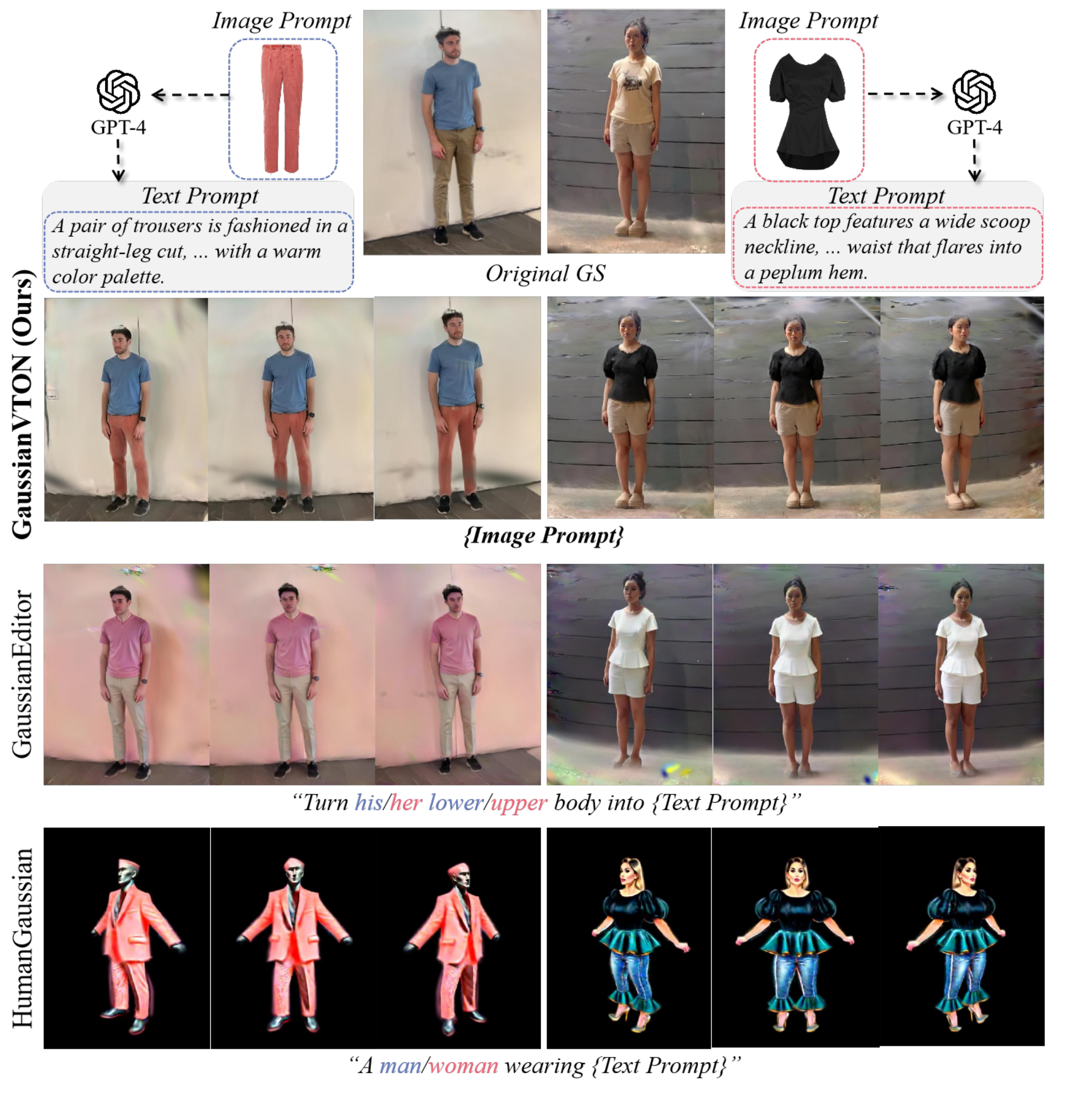}
    \captionof{figure}{\label{fig:teaser}
    GaussianVTON enables efficient human-environment interaction in try-on applications by reconstructing and editing multi-view images. Our method for the first time employs image prompting to achieve more precise and customized 3D Gaussian Splatting editing. Unlike existing works that rely on text prompts for 3D editing, \textit{e.g.}, GaussianEditor~\cite{chen2023gaussianeditor}, our method avoids erroneously replacing clothing and affecting other areas of the garment, as well as causing changes in other elements like background and facial features. Furthermore, compared to text-driven 3D clothed human generation or reconstruction works, \textit{e.g.}, HumanGaussian~\cite{liu2023humangaussian}, our method is based on real human images, avoids resulting in odd body shapes, and aligns with the prompts.
    } 
\end{figure}

The rise of e-commerce and online shopping has led to significant advancements in recommendation systems~\cite{sarkar2022outfittransformer, cucurull2019contextaware, 10.1145/3531017, hsiao2018creating} and visual or keyword-based search~\cite{wu2020fashion, baldrati2023composed, liu2016deepfashion, Morelli2021FashionSearchIC}. From the perspective of potential clothing consumers, accurately identifying their needs is crucial. However, determining the suitability of the clothing before making a purchase decision is equally significant.  Generally, consumers typically rely on product images, model photos, and buyer reviews often leads to uncertainty.
Virtual Try-On (VTON)~\cite{morelli2023ladivton, bai2022single, choi2021vitonhd, 9857232, han2018viton, issenhuth2020mask, morelli2022dress, wang2018characteristicpreserving, Gou_2023}, facilitated by advancements in Generative Adversarial Networks (GANs)~\cite{goodfellow2014generative} and Latent Diffusion Models (LDMs)~\cite{rombach2022high}, allows users to virtually try on garments using catalog images of persons and clothing items.
However, exsiting research~\cite{niu2024pfdm, shim2023squeezingaverse, ning2023picture, zhang2023warpdiffusion} primarily focuses on image-based (2D) VTON and lacks robustness with custom user images~\cite{mohammadi2021smart}. While traditional 3D VTON~\cite{NEURIPS2022_4ee3ac2c, majithia2021robust, jiang2020bcnet, santesteban2021selfsupervised} prioritizes garment design rationality and stylistic effects over authentic human body simulation, which has extensively covered in 2D VTON. These limitations restrict the convenience that VTON currently provides to users.


Recently, the rise of implicit 3D representations, represented by Neural Radiation Fields (NeRF)~\cite{mildenhall2021nerf}, marks a paradigm shift in 3D scene rendering~\cite{li2023spectralnerf, wang2023adaptive, hu2024learning, kaneko2023mimonerf, pan2023transhuman}. This has also spurred research aimed at user-friendly 3D representations and editing algorithms~\cite{haque2023instructnerf2nerf, song2023efficientnerf2nerf, dong2024vicanerf, bartrum2024replaceanything3dtextguided}, which achieve 3D editing by employing a 2D diffusion model to edit each multi-viewpoint. 
Unlike conventional reconstruction or generation~\cite{li2024gaussianbody, lee2023dynamic, dai2023cloth2body, cao2023sesdf, liu2023humangaussian} of 3D clothed humans using meshes or SMPL models~\cite{10.1145/3596711.3596800}, reconstructing 3D scenes using real multi-view human images can yield more realistic results and mitigate issues, \textit{e.g.}, odd proportions, face blurring, etc., as shown in the bottom of~\cref{fig:teaser}. With its real-time rendering and explicit point cloud-like representation, the recent advent of Gaussian Splatting (GS)~\cite{kerbl20233d} further empowers 3D scene editing techniques, rendering them more flexible and rapid.
Whether NeRF or GS-based 3D scene editing, current methods predominantly allow for editing exclusively through text prompts~\cite{haque2023instructnerf2nerf, koo2023posterior, fang2023gaussianeditor, chen2023gaussianeditor}, relying on 2D diffusion models (\textit{e.g.}, InstructPix2Pix~\cite{brooks2023instructpix2pix}) for iterative editing of images from each viewpoint.
However, text-driven editing usually leads to discrepancies between the users' expectations and the actual output, especially in tasks like VTON that require specific editing for certain areas as demonstrated in~\cref{fig:teaser}.

To this end, our work is dedicated to advancing 3D scene editing, specifically by devising a novel VTON pipeline. We diverge from previous methods, which primarily enabled the editing within a scene through text prompts~\cite{haque2023instructnerf2nerf, koo2023posterior, fang2023gaussianeditor}. Instead, we adopt image prompts for personalized editing based on user-specified garment images, thereby achieving a more human-centric experience, given that people tend to focus more on images rather than textual descriptions when shopping online.
However, the development of this work also faces several challenges. A primary issue is the inconsistency in the editing results across different viewpoints when employing a 2D diffusion model for editing multi-view images within a 3D scene~\cite{koo2023posterior, chen2023gaussianeditor}. If edits are guided by text prompts for just clothing color (\textit{e.g.}, \textit{"Turn his pants red."}) or style (\textit{e.g.}, \textit{"Turn him into a clown."}), the diffusion model tends to output somewhat similar results across each viewpoint, though not entirely consistent~\cite{haque2023instructnerf2nerf, fang2023gaussianeditor}. 
Nonetheless, VTON necessitates customized editing of specific regions based on specific images, posing a significant challenge to achieving consistency across multi-view. Additionally, existing 2D VTON models fail to maintain robustness on custom user data~\cite{morelli2023ladivton, bai2022single, choi2021vitonhd, 9857232, han2018viton}. Even within trained datasets, issues including facial blurring and garment inaccuracies arise.

In this work, we propose GaussianVTON, a novel 3D Virtual Try-On framework leveraging Gaussian Splatting editing. GaussianVTON can achieve highly realistic virtual try-ons through 3D scene editing, eliminating the need for additional pre-training data for the 2D VTON model. To ensure optimal multi-view human image editing results, we adopt the state-of-the-art 2D VTON model, LaDI-VTON~\cite{morelli2023ladivton} as the image editing model, and a swift and controllable 3D editing framework, GaussianEditor~\cite{chen2023gaussianeditor} as the 3D editing pipeline. 
To handle the mentioned various issues and challenges, we propose a three-stage refinement strategy:
a) In Stage-1, we devise a simple yet effective facial optimization method aimed at precisely detecting facial keypoints and generating corresponding fusion masks. This addresses the prevalent issue of facial blurring during the editing of 2D VTON models, particularly when handling custom data.
b) In Stage-2, to mitigate the inconsistencies arising from the diffusion model during multi-view editing and to improve the robustness of the 2D VTON model against custom data, we propose \textit{hierarchical sparse editing} to automatically identify views with suboptimal editing quality from all initially edited views for further refinement.
c) In Stage 3, to enhance the quality of editing quality and mitigate potential issues like degradation in viewpoint quality, we draw inspiration from prior works~\cite{koo2023posterior, raj2023dreambooth3d} and employ DDNM~\cite{wang2022zero} to optimize each viewpoint.

Additionally, to address the issues posed by the widely used\cite{haque2023instructnerf2nerf, chen2023gaussianeditor, fang2023gaussianeditor, mirzaei2023watch} editing strategy, \textit{i.e.}, Iterative Dataset Update (IterativeDU), which often leads to complex geometric changes or the addition of objects in unspecified regions during the editing process, we propose a novel strategy termed Editing Recall Reconstruction (ERR). ERR incorporates multi-view images for rendering in a manner consistent with the initial reconstruction process.



To summarize, our contributions are as follows:
\vspace{-0.15cm}
\begin{itemize}
    \item We propose GaussianVTON, a novel 3D Virtual Try-On pipeline leveraging Gaussian Splatting editing via 2D VTON. To the best of our knowledge, GaussianVTON represents the first image prompting 3D editing framework.
    \item We have specifically designed a three-stage refinement strategy aimed at effectively mitigating various disturbances encountered in the process from 2D to 3D editing. Additionally, to better address potential challenges, \textit{e.g.}, complex geometric changes, we propose a novel editing strategy termed Edit Recall Reconstruction (ERR).
    \item Extensive experiments validate the efficacy of GaussianVTON, offering novel insights for both 3D VTON and 3D editing. We believe that our research findings distinctly highlight the potential of employing 3D editing in VTON, aligning more closely with user preferences and thus establishing a promising starting point for future research.
\end{itemize}

\section*{2. Related Work}
\addcontentsline{toc}{section}{Related Work}

\noindent{\bf 2D Virtual Try-On.} Image-based Virtual Try-On (VTON) aims to transfer a desired garment onto the corresponding region of a target subject while preserving human pose and identity. Most existing works~\cite{bai2022single, choi2021vitonhd, chopra2021zflow, dong2019multipose, ge2021parserfree, Han_2019_ICCV, he2022stylebased, issenhuth2020mask, lee2022highresolution, li2021accurate} follow a two-stage generative framework. Some previous works~\cite{choi2021vitonhd, dong2019multipose, issenhuth2020mask, wang2018characteristicpreserving, yang2020photorealistic, 9008110} employ neural networks to regress sparse clothing control points in the target image. These points are then fitted to a Thin Plate Spline (TPS) transformation~\cite{24792} to achieve clothing deformation. Conversely, another research line~\cite{bai2022single, chopra2021zflow, ge2021parserfree, Han_2019_ICCV, he2022stylebased, xie2023gpvton, morelli2023ladivton} estimates an appearance flow map to model non-rigid deformation, where the flow map~\cite{zhou2017view} describes the dense correspondence of each pixel in the target image to its corresponding position in the source image. It's noteworthy that prior works often relied on GANs~\cite{goodfellow2014generative} during the generation stage. However, LaDI-VTON~\cite{morelli2023ladivton} represents the first 2D VTON model entirely based on diffusion models~\cite{rombach2022high}. 
Despite the rapid progress in VTON-related research, their performance on test data still involves issues, \textit{e.g.}, clothing errors, face blurring, etc., not to mention when applied to custom data without pre-training. In this work, we propose a three-stage refinement strategy to achieve better results on custom data editing. Stage-1 addresses problems like face blurring. Stage-2 enhances the consistency of editing through \textit{hierarchical sparse editing}, while Stage-3 mitigates the quality degradation during editing.

\noindent{\bf 3D Virtual Try-On.}
In contrast to 3D human body or clothed human reconstruction~\cite{habermann2020deepcap, gilbert2018volumetric, 9010381, yu2019simulcap, li2024gaussianbody, lee2023dynamic, dai2023cloth2body, cao2023sesdf}, which have been extensively studied, 3D VTON presents unique challenges due to the intricate deformations of clothing. Most research~\cite{NEURIPS2022_4ee3ac2c, wang2018learning, santesteban2019learningbased, patel2020tailornet, bertiche2021pbns, 10.1145/2185520.2185531, saito2020pifuhd, bhatnagar2019multigarment, zhao2021m3dvton, mir2020learning} in this domain focus on estimating digital representations of 3D clothing and understanding how they deform relative to different body shapes. For instance, ULNeF~\cite{NEURIPS2022_4ee3ac2c} introduces a novel neural field-based approach capable of handling multiple garments, while 
MGN~\cite{bhatnagar2019multigarment} predicts parameterized clothing geometry and situates it on the SMPL model~\cite{10.1145/3596711.3596800}, enabling dressing for various body shapes and poses, though constrained to pre-defined digital wardrobes. M3D-VTON~\cite{zhao2021m3dvton} employs edited human images for 3D human reconstructions, however, the reconstructed bodies often suffer from deformations. Additionally, Pix2Surf~\cite{mir2020learning} aims to transfer clothing images to the SMPL model by learning dense correspondences between 2D clothing shapes and 3D clothing UV maps. Existing 3D VTON methods primarily focus on garment fitting but lack consistency with real human data, a gap filled by 2D VTON. To this end, our work pioneers a novel framework for 3D VTON by employing 3D scene editing with image prompting to advance from 2D to 3D VTON.

\noindent{\bf 3D Editing.}
Editing neural fields is challenging due to the complexity of their shape and appearance. EditNeRF~\cite{liu2021editing} is a pioneering work that edits the shape and color of neural fields by adjusting the latent code of the neural field. Additionally, with the application of CLIP~\cite{wang2018learning} across various fields~\cite{yan2023urban, foteinopoulou2023emoclip, You_2023, ma2022xclip, huang2024crest}, some works~\cite{wang2022nerfart, bao2023sine, gao2023textdeformer, wang2022clipnerf} also employ CLIP to facilitate editing the scene reconstructed from source images through text prompts. Another research line~\cite{noguchi2021neural, peng2021neural} focuses on predefined template models or skeletons to support operations, \textit{e.g.} repositioning or rerendering within specific categories. InstructN2N (IN2N) ~\cite{haque2023instructnerf2nerf} proposes a text-driven NeRF editing method termed IterativeDU. It iteratively replaces the reference images originally used for NeRF ~\cite{mildenhall2021nerf} reconstruction with edited images using a 2D diffusion model, InstructP2P (IP2P)~\cite{brooks2023instructpix2pix}. By applying the reconstruction loss of these iteratively updated images to the input 3D scene, the scene gradually transforms into an edited scene. GaussianEditor~\cite{chen2023gaussianeditor} applies Gaussian Splatting~\cite{kerbl20233d} to IN2N, adopting Gaussian semantic tracking to track target Gaussian values, significantly improving editing speed and controllability. 
Additionally, recent works~\cite{li2023focaldreamer, koo2023posterior, park2023ednerf, zhuang2023dreameditor} also apply SDS~\cite{poole2022dreamfusion}, DDS~\cite{hertz2023delta}, and PDS~\cite{koo2023posterior} to optimize IterativeDU. However, these optimizations are all devised for text prompts.  For tasks needing precise personalization like VTON, image prompting is more effective, since text cannot convey all that a garment can be. Thus, we introduce a novel image-prompting 3D editing framework with a new editing strategy termed Editing Recall Reconstruction (ERR).

\section*{3. Method}
\addcontentsline{toc}{section}{Method}

\begin{figure}
    \centering
    \small
   \includegraphics[width=1\linewidth]{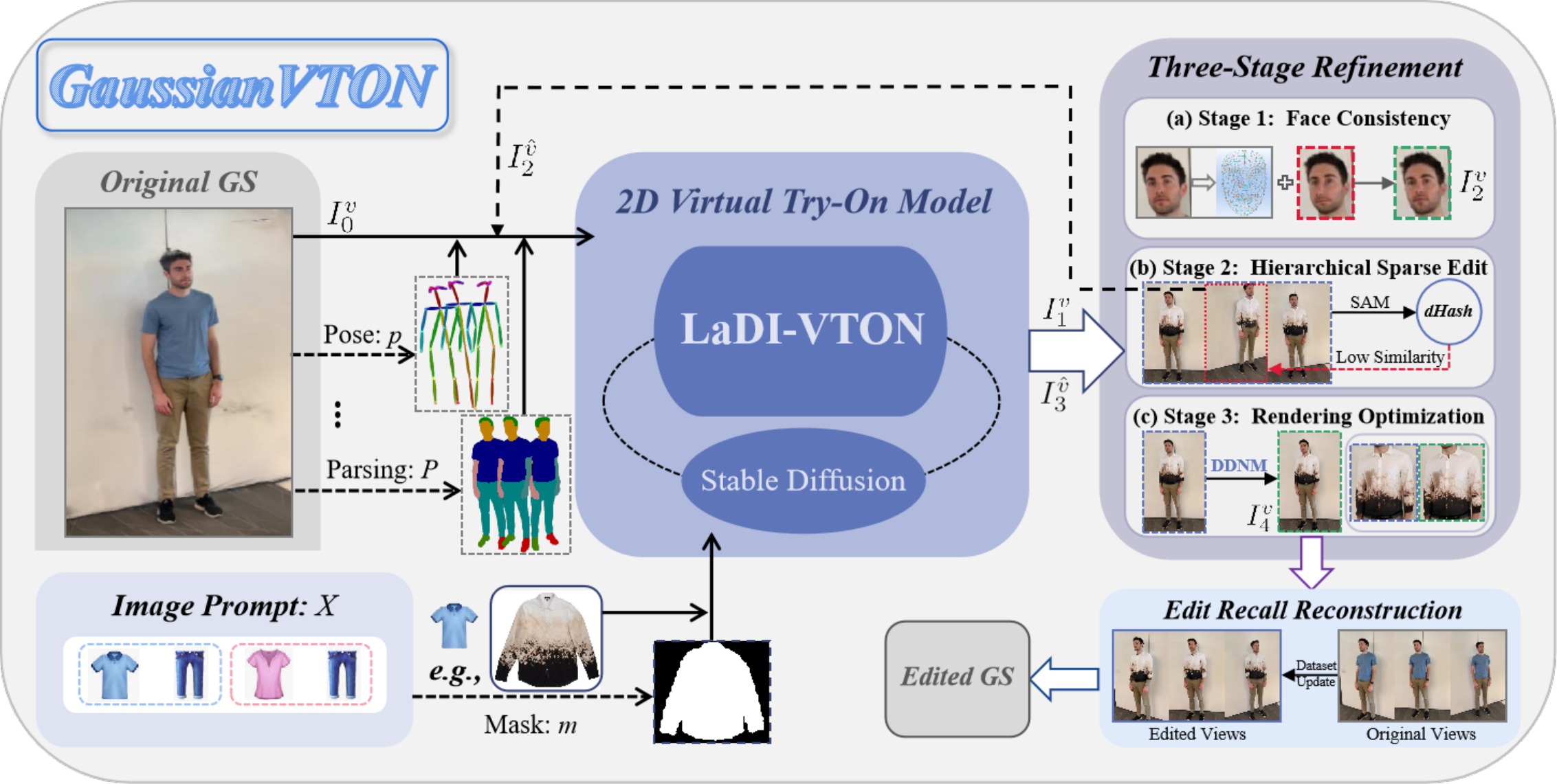}
    \captionof{figure}{\label{fig:framework}
    \textbf{Overall framework of the proposed GaussianVTON.
    }} 
\end{figure}

Our GaussianVTON framework is shown in~\cref{fig:framework}. We start by taking in a reconstructed 3D scene along with its associated data: a series of captured images, their corresponding camera poses, and camera calibration parameters. The main idea revolves around utilizing image prompts to guide the editing process of 3D scenes to achieve virtual try-on. We first introduce our main pipeline, namely the 3D Gaussian Splatting and diffusion-based 2D VTON models in Sec. \textcolor{red}{3.1}. Following that, we present our newly proposed editing strategy, Editing Recall Reconstruction (ERR), along with a novel three-stage refinement strategy in Sec. \textcolor{red}{3.2}.

\subsection*{3.1. Background}
\addcontentsline{toc}{subsection}{Background}

\label{sect:background}
\noindent{\bf 3D Gaussian Splatting.} 
Gaussian Splatting (GS)~\cite{kerbl20233d} embodies a direct depiction of a 3D scene utilizing point clouds, which are differentiable and readily projected to 2D splats, enabling rapid $\alpha$-blending for rendering purposes. Within this, GS characterizes geometry through a collection of 3D Gaussians, obviating the need for normal vectors. These Gaussians are delineated by a complete 3D covariance matrix $\Sigma$, situated in world space~\cite{964490} and centered around point $\mu$:
\begin{equation}  
G(x) = e^{-\frac{1}{2} x^{T} \Sigma^{-1} x}. \label{eq1}
\end{equation}  
where $x$ denotes each Gaussian as a center point.
To render 3D Gaussians, the projection method introduced in~\cite{964490} is employed, which entails a viewing transformation denoted by $W$ and the Jacobian $J$ of the affine approximation of the projective transformation. Employing these elements, the covariance matrix $\Sigma'$ in camera coordinates is computed as follows:
\begin{equation}
    \Sigma'=JW\Sigma W^{T}J^{T}. \label{eq2}
\end{equation}

\noindent The covariance matrix $\Sigma$ is analogous to describing the configuration of an ellipsoid, which can be decomposed into a rotation
matrix \textbf{R} and a scaling matrix \textbf{S} for differentiable optimization (the gradient flow is detailed in \cite{964490}):
\begin{equation}
    \Sigma=\mathbf{RSS}^T\mathbf{R}^T, \label{eq3}
\end{equation}

To summarize, each Gaussian point in the model is characterized by a set of attributes: its position, denoted as $x\in\mathbb{R}^3$, its color represented by spherical harmonic coefficients $c\in\mathbb{R}^k$
(where $k$ indicates the degrees of freedom), its opacity $\alpha\in\mathbb{R}$, a rotation quaternion $q\in\mathbb{R}^4$, and a scaling factor $s\in\mathbb{R}^3$. Specifically, the color $C$ is given by volumetric rendering along a ray with transmittance $T$:
\begin{equation}
    C=\sum_{i=1}^NT_i\alpha_i\mathbf{c}_i, \label{eq4}
\end{equation}
with $\alpha_i=(1-\exp(-\sigma_i\delta_i))$ and $T_i=\prod_{j=1}^{i-1}(1-\alpha_i).$
Specifically, the color and opacity of each Gaussian are computed for every pixel according to the Gaussian representation outlined in \cref{eq1}. The blending mechanism for $N$ ordered points overlapping a pixel adheres to a prescribed formula:
\begin{equation}
    C=\sum_{i\in N}c_i\alpha_i\prod_{j=1}^{i-1}(1-\alpha_j). \label{eq5}
\end{equation}
where $c_i$ and $\alpha_i$ signify the color and density of a given
point respectively. 
A Gaussian distribution calculates these values using a covariance matrix $\Sigma$, which is subsequently adjusted by per-point opacity and spherical harmonics (SH) color coefficients that are subject to optimization.

\noindent{\bf 2D Try-On Diffusion Model.} In recent research, numerous endeavors have elevated 2D diffusion processes to 3D editing. One notable focus is multi-view rendering based on a given 3D model, enabling 2D editing through text prompts~\cite{chen2023it3d, brooks2023instructpix2pix, raj2023dreambooth3d}.
This involves generating a dataset of multi-view images to train and guide the 3D models. Our work aims to exploit advanced 3D scene editing techniques to explore the potential for achieving more realistic and personalized 3D VTON via 2D VTON, constituting the first exploration for image-prompting 3D editing. Hence, we have not designed a specific 2D VTON method, implying that any diffusion-based 2D VTON model can be applied in our framework. 

To facilitate comprehension, we briefly introduce the 2D VTON model adopted in this work, termed LaDI-VTON~\cite{morelli2023ladivton}, which builds upon a latent diffusion model~\cite{rombach2022highresolution} and extends it with an additional autoencoder module. This model harnesses learnable skip connections to augment the generation process while preserving key features. 
Given an in-shop garment image $X$, a target model image $I$, and auxiliary inputs including pose keypoints $p$, human parsing $P$, inpainting mask $m$, and dense pose $d$. $X$ undergoes deformation to align with target model pose $p$, resulting in warped image $X_W$. $I_M$ is derived by masking $I$ with $X_W$. This setup is common in 2D VTON models.

LaDI-VTON is designed to generate a new image $\tilde{I}$ by substituting a target garment in $I$ with $X$ provided by the user, while preserving the model’s physical characteristics, pose, and identity. This approach is rooted in the Stable Diffusion inpainting pipeline~\cite{rombach2022highresolution}. The spatial input $\gamma$ comprises the channel-wise concatenation of an encoded masked image $\mathcal{E}(I_M)$, a resized binary inpainting mask $m\in\{0,1\}^{1\times h\times w}$, and input from the denoising network $z_t$. Specifically, $I_M$ denotes $I$ masked according to the inpainting mask $M\in\{0,1\}^{1\times H\times W}$, and $m$ is resized based on the spatial dimensions of the original mask $M$. Thus, the spatial input for the inpainting denoising network is as follows:
\begin{equation}
    \gamma=[z_{t};m;{\mathcal E}(I_{M})]\in\mathbb{R}^{(4+1+4)\times h\times w}. \label{eq6}
\end{equation}

In summary, LaDI-VTON relies on a latent diffusion model~\cite{rombach2022highresolution}, where the variables consist of latent images generated by encoding an RGB image. Likewise, to generate an RGB image from the diffusion model, it is necessary to decode the predicted $\tilde{z}_0$ latent variables using the decoder $\tilde{I}={\mathcal D}(\tilde{z}_{0})$.

\subsection*{3.2. GaussianVTON}
\addcontentsline{toc}{subsection}{GaussianVTON}
\label{sect:gaussianvton}

Given a reconstructed 3D GS scene along with the corresponding dataset of calibrated images $I^v$ with view index $v$, and an image prompt of the target garment $X$ for try-on, we employ fine-tuning to adjust the reconstructed scene based on editing instructions. This process results in an edited version of the GS scene, which is presented within the framework as illustrated in \cref{fig:framework}.

In this work, we employ LaDI-VTON~\cite{morelli2023ladivton} to edit each image $I^v$ within the dataset. 
As mentioned in Sec. \textcolor{red}{3.1}, 2D VTON requires various inputs, some of which cannot be directly obtained from everyday life.
Such intricate input requirements also pose challenges when utilizing custom data. Hence, to ensure the user-friendly purpose of 3D editing, we devised a solution within GaussianVTON to automatically extract relevant input data, thus necessitating only the provision of the target model image $I$ and the target garment image $X$ as inputs (more details are provided in Sec. \textcolor{red}{4.1}). In the upcoming part, we present our newly proposed editing strategy termed Editing Recall Reconstruction (ERR) and the three-stage refinement strategy aimed at effectively mitigating various issues encountered during the editing process, thereby facilitating better integration between 2D VTON models and 3D editing methods.

\noindent{\bf Edit Recall Reconstruction.}
Edit Recall Reconstruction (ERR) refers to the rendering of the entire dataset during editing at the same time, in a similar manner as the reconstruction process. At the onset of optimization, our image dataset comprises the original images $I_{0}^{v}$, where $0$ denotes the editing step remains 0. These images are cached independently and employed as conditioning inputs for the diffusion model throughout all stages. Subsequently, ERR proceeds sequentially according to the viewpoints $v$, using the target garment $X$ as the prompt to perform editing on all $I_{0}^{v}$ images via LaDI-VTON. Upon completion of editing and refinement for the entire dataset, the dataset update occurs, followed by rendering for this iteration. 

Differs from the editing strategy, IterativeDU~\cite{haque2023instructnerf2nerf}, commonly used in previous studies~\cite{haque2023instructnerf2nerf,chen2023gaussianeditor,dong2024vicanerf}, which iteratively replaces and updates dataset images. ERR waits until all images in the dataset have been edited and refined before updating the dataset, effectively alleviating inconsistent editing issues, \textit{e.g.}, complex geometric alterations and the addition of objects in unspecified regions caused by inconsistent editing moments (further analysis in Sec. \textcolor{red}{4.4}). Meanwhile, compared to recent applications, \textit{e.g.}, SDS~\cite{poole2022dreamfusion}, DDS \cite{hertz2023delta} and PDS~\cite{koo2023posterior}, ERR demonstrates the ability to accommodate image editing prompts without inducing fundamental changes in the edited views compared to the input views.


\noindent{\bf Three-Stage Refinement.}
As previously discussed, both 2D editing and 3D editing suffer from various drawbacks such as multi-view inconsistency, partial blurring (\textit{e.g.} face blurring), and degradation of image quality due to diffusion models~\cite{rombach2022highresolution, brooks2023instructpix2pix, morelli2023ladivton}. To mitigate these issues, GaussainVTON adopts a three-stage refinement strategy outlined as follows (specific examples are shown in \cref{fig:3-stage}):

\begin{figure}[t]
    \centering
    \small
   \includegraphics[width=1\linewidth]{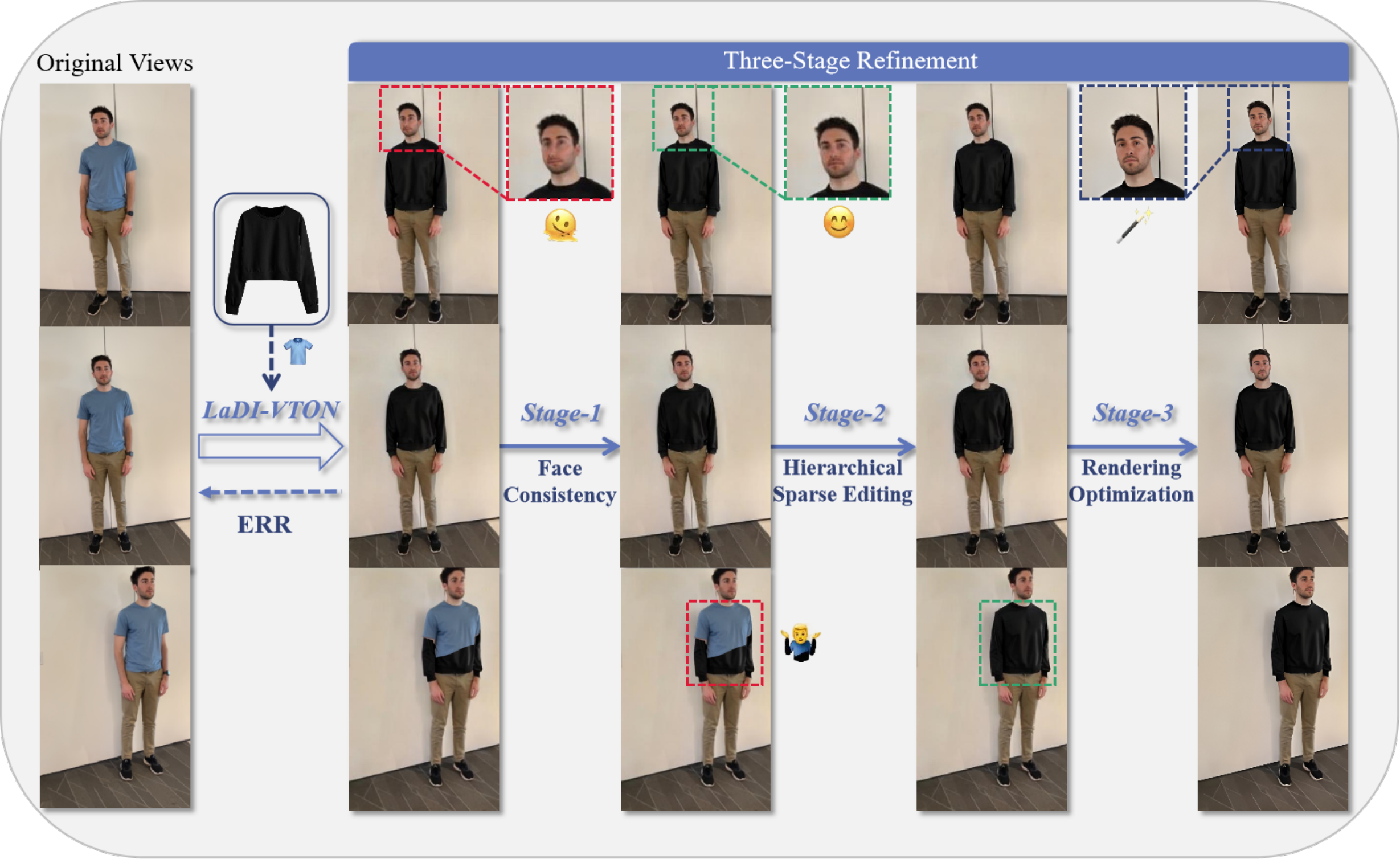}
    \captionof{figure}{\label{fig:3-stage}
    \textbf{Three-Stage Refinement.} The three-stage refinement strategy we devised demonstrates sequential mitigation of prominent issues encountered when utilizing 2D VTON models (\textit{i.e.} LaDI-VTON) for image editing, including facial blurring, garment inaccuracies, and degradation in image quality.
    }
\end{figure}

\noindent {\textit{a) Stage-1: Face Consistency.}}
Face blurring is a pervasive yet significant issue encountered in various computer vision tasks, including VTON tasks~\cite{morelli2023ladivton, bai2022single, choi2021vitonhd, chopra2021zflow, dong2019multipose, zhao2021m3dvton}, pose transfer tasks~\cite{cheong2023upgpt, bhunia2023person, zhang2021pise, zhou2022cross}, and 2D/3D human reconstruction or generation tasks~\cite{jiang2022text2human, yu2019simulcap, li2024gaussianbody, lee2023dynamic, dai2023cloth2body, cao2023sesdf}. 
However, in tasks like 3D scene reconstruction, facial features within the scene are not heavily affected~\cite{haque2023instructnerf2nerf, chen2023gaussianeditor, dong2024vicanerf, koo2023posterior, fang2023gaussianeditor}. Thus, to address this issue in GaussianVTON, which is mainly caused by VTON, we first employ the FaceMesh model from MediaPipe~\cite{lugaresi2019mediapipe} to initialize a facial keypoint detector $K(\cdot)$. For each input view image $I_{0}^{v}$, upon detecting facial information within this view, facial keypoints are extracted and stored as the facial network $F^{v}=K(I_{0}^{v})$. Subsequently, after the view $I_{0}^{v}$ undergoes editing via LaDI-VTON, resulting in ${I}_{1}^{v}$, a matching process is employed between the pre- and post-edited views according to the index of view $v$. Finally, the extracted facial network $F^{v}$ is transferred to the post-edited view ${I}_{1}^{v}$, where facial distortion is more likely to occur.

\noindent {\textit{b) Stage-2: Hierarchical Sparse Editing.}}
In addition to face blurring, the inconsistency in editing between multi-views is also a major challenge faced by GaussianVTON during the editing process. Previous works~\cite {koo2023posterior,haque2023instructnerf2nerf} have confirmed that this issue arises primarily due to the inability of diffusion models~\cite{rombach2022highresolution} to achieve consistency. However, not only that, VTON typically relies on a large amount of training data to achieve relatively satisfactory results on relevant data, but issues like clothing errors still exist. Therefore, to mitigate the impact of the aforementioned problems, especially on custom data for GaussianVTON, we propose \textit{hierarchical sparse editing}.

In the context of post-editing view images after face refinement, denoted as ${I}_{2}^{v}$, we employ Lang-Segment-Anything based on Segment Anything (SAM)~\cite{kirillov2023segment} as our segmentation model $S(\cdot)$ to obtain masks $M_{W}^{v}=S({I}_{2}^{v})$ corresponding to the edited garment regions (\textit{e.g.}, upper clothing, lower clothing). Subsequently, we retrieve the corresponding images $I_{M_W}^{v}$ based on these masks. Sequentially, following the order of view $v$, we acquire every three segmented garment images $I_{M_W}^{i}$, $I_{M_W}^{j}$, $I_{M_W}^{k}$ associated with view indices $v=i$, $v=j$, $v=k$ in order, respectively. We then employ the \textit{dHash} algorithm to compute the similarity between these three images, and identify the view index $\hat{v}$ of the image with the lowest similarity to the other two images:
\begin{equation}
    \begin{aligned}
    &\hat{v} = \min_{i\neq j\neq k}(\max(dHash(I_{M_W}^{i},I_{M_W}^{j}),
    \\
    &dHash(I_{M_W}^{i},I_{M_W}^{k}),dHash(I_{M_W}^{j},I_{M_W}^{k}))). \label{eq7}
    \end{aligned}
\end{equation}
This approach aids in detecting potential errors in the edited garment, providing an approximate assessment of garment fidelity.
After traversing all segmented garment images $I_{M_W}$ in the dataset, indices $\hat{v}$ corresponding to all low-similarity images are obtained. Subsequently, the perspective image ${I}_{2}^{\hat{v}}$, corresponding to the post-face refinement one, is re-edited with the target garment $X$ as the image prompt to achieve the refinement. 

\noindent {\textit{c) Stage-3: Rendering Optimization.}} Given the possibility of image quality deterioration or distortion during editing, we are inspired by previous research \cite{raj2023dreambooth3d, koo2023posterior} which opt to employ SDEdit \cite{meng2022sdedit} to denoise images. In this work, we adopt DDNM~\cite{wang2022zero}, which shows promising performance in image restoration via denoising diffusion null-space model:
\begin{equation}
\mathbf{y}=\mathbf{A}\mathbf{x}+\mathbf{n},\quad\mathbf{n}\in\mathbb{R}^{d\times1}\sim\mathcal{N}(\mathbf{0},\sigma_\mathbf{y}^2\mathbf{I}), \label{eq8}
\end{equation}
where $\mathbf{x}\in\mathbb{R}^{D\times1}$ represents the original image, $\mathbf{A}\in\mathbb{R}^{d\times D}$ denotes the degradation matrix, and $\mathbf{y}\in\mathbb{R}^{d\times1}$ represents the degraded image.
Through this approach, it becomes feasible to refine the rendering $\tilde{I}_{3}^{\hat{v}}$ of the image after re-editing and $\tilde{I}_{2}^{v-\hat{v}}$ that has not undergone hierarchical sparse editing, into a more visually realistic representation.
This ultimately yields the final perspective views ${I}_{4}^{v}$ to update the dataset.

\section*{4. Experiments}
\addcontentsline{toc}{section}{Experiments}

\label{sec:experiments}

\begin{figure*}[t]
    \centering
   \includegraphics[width=1\linewidth]{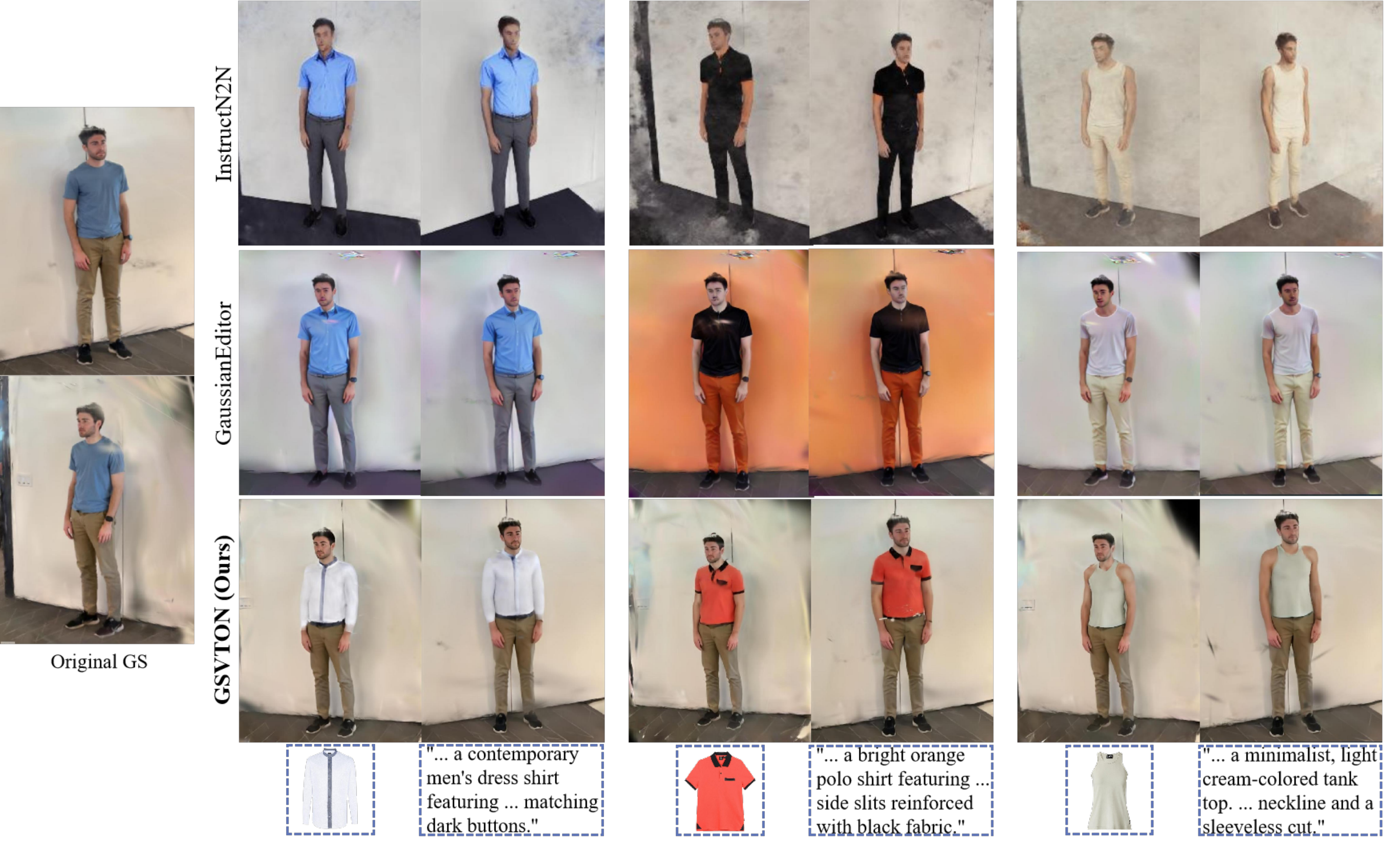}
    \captionof{figure}{\label{fig:qualitative}
    \textbf{Qualitative Comparison.} We ask GPT-4~\cite{openai2023gpt4} to generate detailed descriptions of target garments, followed by format \textit{"Turn his upper body into ..."} as the text prompt for InstructN2N~\cite{haque2023instructnerf2nerf} and GaussianEditor~\cite{chen2023gaussianeditor}. We adopt GSEditor-iN2N from GaussianEditor as the comparative model due to its superior performance.
    }
\end{figure*}

\subsection*{4.1. Implementation Details}
\addcontentsline{toc}{subsection}{Implementation Details}

\label{sect:implementation}
We adopt the highly optimized renderer implementation from Gaussian Splatting~\cite{kerbl20233d} for Gaussian rendering, as proposed in~\cite{chen2023gaussianeditor}.
For the input multi-view human data, the original 3D Gaussians employed in this work are trained using the method outlined in~\cite{kerbl20233d}.
Subsequently, we employ the 2D VTON model, LaDI-VTON~\cite{morelli2023ladivton}, pre-trained on DressCode~\cite{morelli2022dress}, without further training on the multi-view data we use. To simplify the inputs, we only use the multi-view human images as target model images and target garment images as image prompting. GaussianVTON automatically preprocesses these data, including obtaining the pose keypoints of the target model using OpenPose~\cite{cao2019openpose}, extracting the human parsing using~\cite{li2020self}, and obtaining the target human dense pose with Detectron2~\cite{wu2019detectron2}. We commit to releasing all the data and code of this work. Additionally, we mostly adhere to the hyperparameter settings outlined in GaussianEditor~\cite{chen2023gaussianeditor} and LaDI-VTON~\cite{morelli2023ladivton}. Moreover, previous works~\cite{chen2023gaussianeditor, haque2023instructnerf2nerf} enhance the editing performance by cropping multi-view images to a limited size, \textit{e.g.} 512×512.
To ensure the overall editing effect and controllable area of custom data remains unaffected, we employ adaptive sizing editing on all input data.

\subsection*{4.2. Qualitative Comparisons}
\addcontentsline{toc}{subsection}{Qualitative Comparisons}
\label{qualitative}

\begin{figure*}[t]
    \centering
    \small
   \includegraphics[width=1\linewidth]{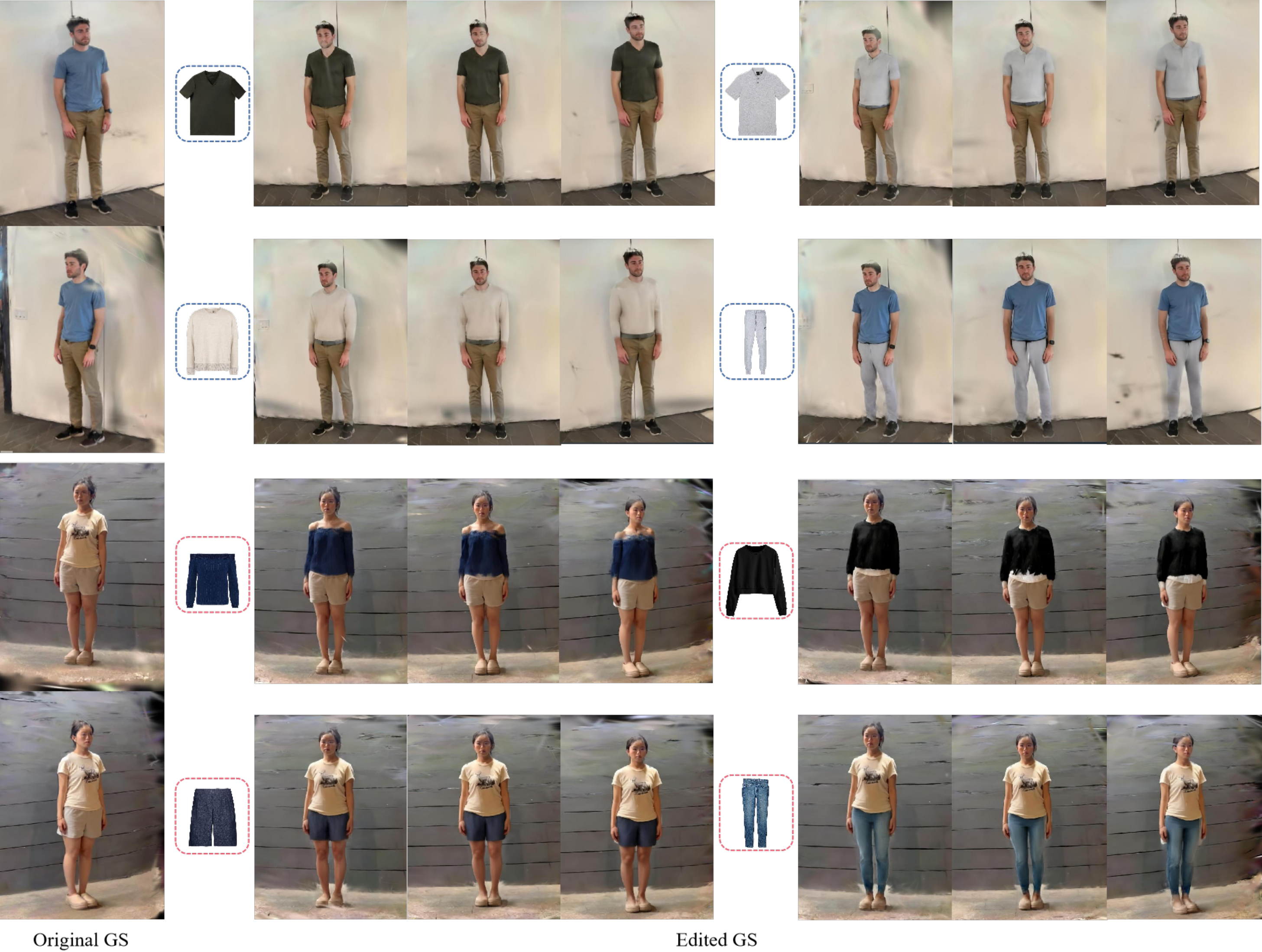}
    \captionof{figure}{\label{fig:results}
    \textbf{Extensive Results of GaussianVTON.} To further validate the efficacy of our framework, we also employ multi-view image data of a female, which further substantiates the superiority and capability to adopt custom data of GaussianVTON.
    }
\end{figure*}

As discussed earlier, existing works on 3D scene editing \cite{haque2023instructnerf2nerf, chen2023gaussianeditor} rely solely on text prompts for editing (\textit{e.g.}, \textit{"Turn his short sleeves black."}) with InstructPix2Pix~\cite{brooks2023instructpix2pix}. Therefore, for a fairer comparison, given a garment image, we first adopt Multimodal Large Language Model (MLLM), i.e. GPT-4~\cite{openai2023gpt4} to describe the garment, and then input the corresponding description in a specific format (\textit{e.g.}, \textit{"Turn his upper body into \{Garment Description\}."}) into text-driven 3D editing models \cite{haque2023instructnerf2nerf, chen2023gaussianeditor}. More details can be found in the Supplementary Material.

As depicted in \cref{fig:qualitative}, although we describe our desired garments as comprehensively as possible through the use of GPT-4~\cite{openai2023gpt4}, methods~\cite{haque2023instructnerf2nerf, chen2023gaussianeditor} edited based on text prompts fail to meet expectations. Not only do they inaccurately represent clothing in corresponding areas, but they also alter garments in other regions (\textit{e.g.}, upper body). Additionally, editing results in unintended changes in the overall scenario (\textit{e.g.}, background), facial features, and view quality (especially InstructN2N~\cite{haque2023instructnerf2nerf}). However, our GaussianVTON directly employs garment images as image prompts, which enables more personalized editing of humans in 3D scenes compared to methods utilizing text prompts. This further underscores the limitations of text-driven 3D editing when confronted with tasks like Virtual Try-On, which necessitate specific edits in particular regions.
To further demonstrate the superiority of our novel 3D VTON framework, GaussianVTON, in this task, we demonstrate additional editing results in \cref{fig:results}.

\subsection*{4.3. Quantitative Comparisons}
\addcontentsline{toc}{subsection}{Quantitative Comparisons}
\label{quantitative}
Following \cite{haque2023instructnerf2nerf, gal2021stylegannada, ruiz2023dreambooth}, we first present a quantitative comparison of CLIP Text-Image Directional Similarity (Text-Image) and CLIP Image-Image Directional Similarity (Image-Image). Similarly, to ensure a fair comparison, we use corresponding garment descriptions instead of image prompts to compute Text-Image. Subsequently, inspired by the generation tasks\cite{liu2023humangaussian, woo2023harmonyview}, we further employ several evaluation metrics for image quality (\textit{i.e.}, FID, SSIM, PSNR and LPIPS) to evaluate the edited multi-views. Adhering to the principle of human-centricity, we also conduct human evaluations on the editing results. As demonstrated in \cref{tab:quantitative}, GaussianVTON depicts superior performance across all evaluation metrics, which implies that our approach not only ensures relatively precise editing but also effectively manages to mitigate issues like image distortion. Notably, our editing results exhibit a substantial preference over the baseline methods in human evaluation. More details are available in the Supplementary Material.

\begin{table*}[t]
\caption{\textbf{Quantitative Comparisons.} We compare against the text-driven 3D scene editing techniques \textit{i.e.}, InstructN2N~\cite{haque2023instructnerf2nerf} and GaussianEditor-iN2N~\cite{chen2023gaussianeditor}.}
  \centering
  \begin{tabular}{@{}lccccccc@{}}
    \toprule
     &Text-Image$\uparrow$ & Image-Image$\uparrow$ & FID$\downarrow$ & SSIM$\uparrow$ & PSNR$\uparrow$ & LPIPS$\downarrow$ & User Study$\uparrow$ \\ 
    \midrule
    InstructNeRF2NeRF~\cite{haque2023instructnerf2nerf} &  0.1600 & 0.6879 & 295.8 & 0.7321 & 14.02 & 0.3110 & 15.70\%\\
    GaussianEditor-iN2N~\cite{chen2023gaussianeditor} & 0.2071 & 0.7558  & 195.4 & 0.8091 & 16.61 & 0.1686 & 10.68\% \\
    \midrule
    GaussianVTON (Ours) & \textbf{0.3293} & \textbf{0.8481} & \textbf{176.1}  & \textbf{0.8171} & \textbf{18.00} & \textbf{0.1654} & \textbf{73.62\%}\\
    \bottomrule
  \end{tabular}
  \label{tab:quantitative}
  \end{table*}

\begin{table*}[t]
  \caption{\textbf{Ablation Studies.} We adopt the same metrics in \cref{tab:quantitative} to further evaluate the effect of our proposed ERR and the three-stage refinement strategy.}
  \centering
  \begin{tabular}{@{}lcccccc@{}}
    \toprule
    &Text-Image$\uparrow$ & Image-Image$\uparrow$ & FID$\downarrow$ & SSIM$\uparrow$ & PSNR$\uparrow$ & LPIPS$\downarrow$ \\ 
    \midrule
    IterativeDU & 0.2876 & 0.8156 & 197.3 & 0.7866 & 17.52 & 0.2175\\
    ERR & \textbf{0.3293} & \textbf{0.8481} & \textbf{176.1} & \textbf{0.8171} & \textbf{18.00} & \textbf{0.1654}\\
    \midrule
    w/o Three-Stage & 0.2151 & 0.7621 & 142.1 & 0.8407 & 17.95 & 0.1697\\
    After Stage-1 & 0.2481 & 0.7829 & \textbf{136.8} & \textbf{0.8443} & \textbf{18.25} & \textbf{0.1563}\\
    After Stage-2 & 0.2713 & 0.8294 & 162.6& 0.8281 & 18.12 & 0.1695\\
    After Stage-3 & \textbf{0.3293} & \textbf{0.8481} & 176.1 & 0.8171 & 18.00& 0.1654\\
    \bottomrule
  \end{tabular}
  \label{tab:ablation}
  \end{table*}

\subsection*{4.4. Ablation Study}
\addcontentsline{toc}{subsection}{Ablation Study}
\label{sec:ablation}

\begin{figure*}[t]
    \centering
    \small
   \includegraphics[width=0.8\linewidth]{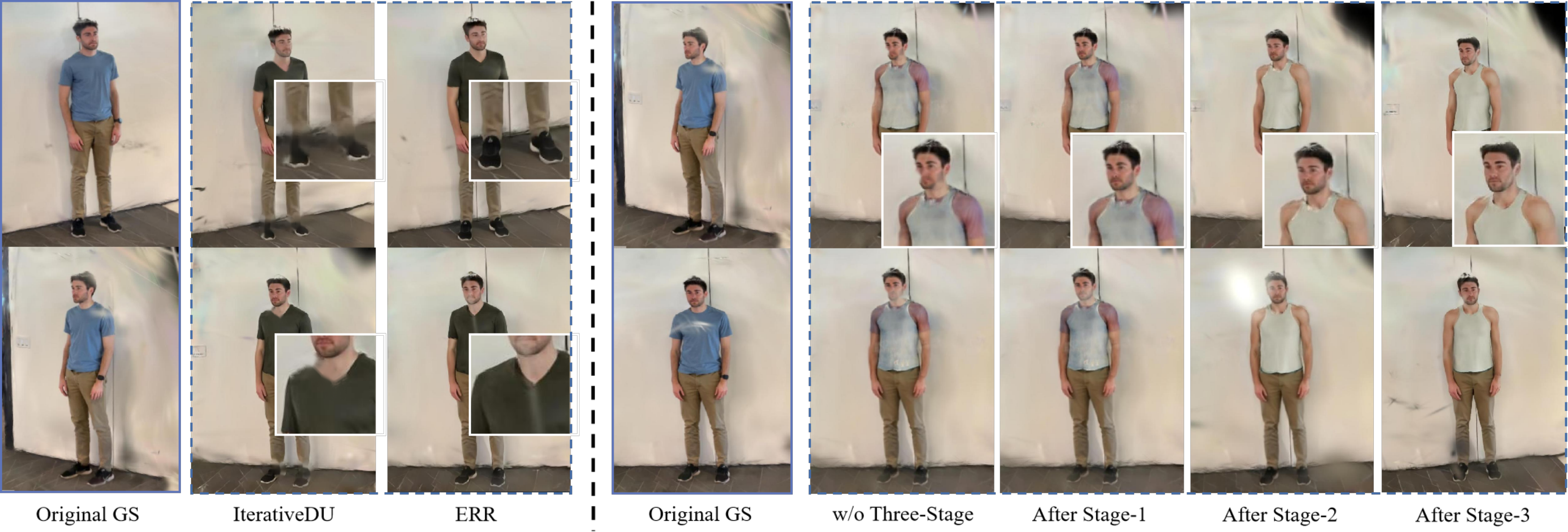}
    \captionof{figure}{\label{fig:ablation}
    \textbf{Ablation Studies.} 
    This figure depicts the results of our proposed editing strategy, ERR, compared to IterativeDU from IN2N~\cite{haque2023instructnerf2nerf} (shown on the \textbf{left}). On the \textbf{right}, the rendering results of our Three-Stage Refinement process are showcased: Stage-1 maintains facial consistency, Stage-2 refines garments using hierarchical sparse editing, and Stage-3 optimizes overall image quality.
    }
\end{figure*}

As demonstrated in the left panel of \cref{fig:ablation}, we further compare our newly proposed editing strategy, namely ERR, with IterativeDU (which is widely used in the text-driven 3D editing framework). It can be observed that, compared to IterativeDU, ERR effectively mitigates complex geometric changes, \textit{e.g.}, missing or deformed regions in garments, by simultaneously updating and rendering, rather than updating the dataset sequentially from each viewpoint.



Furthermore, as illustrated in \cref{fig:3-stage}, we have exemplified specific instance views employing the three-stage refinement strategy. To gain a deeper insight into the impact of potential challenges encountered during the transition from 2D to 3D editing, we conduct ablation experiments on the rendering results of this strategy, depicted in the right panel of \cref{fig:ablation}, which can gradually address various issues, \textit{e.g.}, face blurring, garment inaccuracies and quality degradation.

To provide quantitative analysis, we utilize the same evaluation metrics as presented in \cref{tab:quantitative} to further scrutinize our proposed strategies, as demonstrated in \cref{tab:ablation}. It is noteworthy that following Stage-1 (face consistency), the metrics at the pixel level (\textit{i.e.}, FID, SSIM, PSNR, and LPIPS) reach their optimal values for multi-viewpoints, indicating the achievement of facial optimization to maintain consistency with the original view. However, garment optimization (Stage-2) has not yet been performed, thus preserving more elements of the original attire, as depicted in the right panel of \cref{fig:ablation}. Building upon both qualitative and quantitative ablation analyses, we further substantiate that both ERR and the three-stage refinement strategy underscore the effectiveness of our GaussianVTON framework. This effectiveness is attributed to its innovative optimization designs, which facilitate direct image-prompting 3D editing.



\subsection*{4.5. Limitations}
\addcontentsline{toc}{subsection}{Limitations}

\label{limitation}
Although GaussianVTON shows decent performance, it inherits limitations of the 2D try-on diffusion model within 3D editing. Similar to most existing 2D VTON models, the 2D try-on diffusion model confronts challenges due to its reliance on extensive training data and narrow testing datasets, hindering practical application. This situation poses unique challenges for developing 3D VTON systems tailored to custom user data, as explored in this work. 
Moreover, while GaussianVTON also centers on image-guided 3D editing, VTON cannot encompass its entirety. Future work will advance this framework, contributing to more comprehensive and personalized image-prompting 3D editing.

\section*{5. Conclusion}
\addcontentsline{toc}{section}{Conclusion}
In this work, we propose GaussianVTON, a novel 3D Virtual Try-On (VTON) pipeline leveraging 3D Gaussian Splatting editing, which represents a significant advancement in both image-prompting 3D editing and 3D VTON. Our method enables realistic try-on experiences for users through the reconstruction and editing of real scenes. To address the challenges inherent in transitioning from 2D to 3D editing, our method employs a three-stage refinement strategy. Furthermore, we introduce a specialized editing strategy termed Edit Recall Reconstruction (ERR), which enhances rendering smoothness and prevents undesirable artifacts resulting from complex geometry alterations.
{
    \small
    \bibliographystyle{ieeenat_fullname}
    \bibliography{main}
}

\end{document}